\documentclass[runningheads, anonymous, UTF8]{llncs}

\usepackage[T1]{fontenc}
\usepackage{graphicx}
\usepackage{enumitem}
\usepackage{marvosym}
\usepackage{lineno}
\usepackage{amssymb}
\usepackage{amsmath}
\usepackage{booktabs}
\usepackage{multirow}
\usepackage{multicol}
\usepackage{ulem}
\usepackage{subfigure}
\usepackage{cite}
 
\begin{document}

\title{Integrating Multi-view Analysis: \\Multi-view Mixture-of-Expert for \\ Textual Personality Detection}
\titlerunning{Multi-view Mixture-of-Expert for Textual Personality Detection}
\author{Haohao Zhu
\and Xiaokun Zhang
\and Junyu Lu
\and Liang Yang
\and Hongfei Lin\textsuperscript{(\Letter)}
}

\authorrunning{H. Zhu et al.}

\institute{
School of Computer Science and Technology, Dalian University of Technology, Dalian, China 
\\
\email{zhuhh@mail.dlut.edu.cn, dawnkun1993@gmail.com, ,dutljy@mail.dlut.edu.cn, \{liang,hflin\}@dlut.edu.cn}
}

\maketitle              

\begin{abstract}
Textual personality detection aims to identify personality traits by analyzing user-generated content.
To achieve this effectively, it is essential to thoroughly examine user-generated content from various perspectives.
However, previous studies have struggled with automatically extracting and effectively integrating information from multiple perspectives, thereby limiting their performance on personality detection.
To address these challenges, we propose the \textbf{M}ulti-\textbf{v}iew Mixture-of-Experts Model for Textual \textbf{P}ersonality Detection (MvP).
MvP introduces a Multi-view Mixture-of-Experts (MoE) network to automatically analyze user posts from various perspectives.
Additionally, it employs User Consistency Regularization to mitigate conflicts among different perspectives and learn a multi-view generic user representation.
The model's training is optimized via a multi-task joint learning strategy that balances supervised personality detection with self-supervised user consistency constraints.
Experimental results on two widely-used personality detection datasets demonstrate the effectiveness of the MvP model and the benefits of automatically analyzing user posts from diverse perspectives for textual personality detection.

\end{abstract}

\section{Introduction}

Understanding users' personality information contributes to comprehending their decision-making processes \cite{mendes2019relationship}, behavioral patterns \cite{greeno1973personality}, preference information \cite{furnham1981personality}, making it valuable for various downstream tasks such as recommender systems \cite{dhelim2022survey, zhang2023bi ,zhang2023beyond, zhang2024disentangling}. 
However, traditional personality assessment relies on users' self-assessment reports or face-to-face experiments, which are time-consuming, labor-intensive, and cannot be automated on a large scale \cite{fang2023text}. 
With the prevalence of the internet and social media, the vast amount of information shared by users on social media platforms has made automated personality analysis possible \cite{fang2023text}.

In this context, textual personality detection\footnote{Throughout this paper, unless specifically stated otherwise, personality detection refers to textual personality detection.} is proposed to analyze users' personalities based on the posts they share on social media platforms and has garnered widespread attention\cite{lynn2020hierarchical,yang2023orders,fang2023text}, and has been demonstrated to be helpful in various downstream tasks such as personalized recommendation systems\cite{dhelim2022survey}, personalized advertising\cite{roffo2016personality}, dialog systems \cite{yang2021improving}. 

Personality detection is typically delineated across various dimensions. For instance, in the Myers-Briggs Type Indicator (MBTI) taxonomy \cite{myers1987introduction}, personality detection entails assessing a user's inclination in 4 dimensions, with each dimension emphasizing different behavioral patterns:
E/I (Extraversion/Introversion) pertains to how individuals derive energy and where they focus their attention.
S/N (Sensing/Intuition) relates to how individuals absorb information and their perception style.
T/F (Thinking/Feeling) delves into how individuals make decisions and process information.
J/P (Judging/Perceiving) centers on individuals' approach to the external world and their decision-making style.

Given that various personality traits highlight different behavioral patterns, it's crucial to effectively analyze a user's posts from multiple perspectives.
However, existing personality detection methods often overlook this multifaceted nature of personality by relying on representations learned from a single viewpoint. 
To fill this gap, we introduce the \textbf{M}ulti-\textbf{v}iew Mixture-of-Expert for Textual \textbf{P}ersonality Detection (MvP), designed to automatically analyze user posts from various perspectives within a unified framework. 
MvP leverages a Multi-view Mixture-of-Expert (MoE) to model user posts from different perspectives, thus capturing a comprehensive representation of user content and enabling more accurate personality detection.

However, analyzing user posts from different perspectives simultaneously introduces the challenge of potential conflicts arising among the representations learned from different perspectives, leading to a seesaw phenomenon\footnote{Learning from multiple perspectives is conflict due to their potential interference and redundancies between them.} in personality detection. 
To address this challenge, MvP incorporates User Consistency Regularization, a novel technique that mitigates conflicts between different perspectives. 
User Consistency Regularization enables the integration of diverse representations into a unified user representation, ensuring a more coherent and accurate personality assessment.

The principal contributions of this paper are outlined below:
\begin{enumerate}[label={(\arabic*)}]
	\item We explore how automatically modeling and analyzing user posts from various perspectives impact the personality detection process. Additionally, we explore methods to effectively alleviate conflicts among these perspectives as much as possible.
	\item We propose the MvP framework, employing the Mixture-of-Expert (MoE) to analyze user posts from multiple perspectives, enriching user representation. MvP also integrates user consistency regularization to address conflicts from multi-view representations, ensuring a cohesive assessment of user personalities.
	\item Extensive experiments on benchmark datasets demonstrate the efficacy of the MvP model, highlighting the advantages of multi-view modeling in personality detection. We also analyze the contributions of each key module within MvP, elucidating their roles in enhancing detection performance.
\end{enumerate}

\section{Related Work}

Initially, researchers in personality detection focused on exploring various psychological linguistic features and statistical features, leading to the discovery of feature sets such as LIWC \cite{pennebaker2001linguistic}, Mairesse \cite{mairesse2007using}, SPLI \cite{moffitt2010structured}, and others \cite{cambria2018senticnet, mohammad2013crowdsourcing, mohammad2018obtaining}. 
However, manually extracting post features is time-consuming, laborious, and often misses valuable insights\cite{fang2023text}.
In recent years, methods based on deep learning and pre-trained language models have dominated the field of personality detection due to their ability to avoid manually finding features.
Various neural network architectures such as CNNs \cite{majumder2017deep, tandera2017personality}, RNNs \cite{tandera2017personality, tandera2017personality, lynn2020hierarchical}, attention mechanisms \cite{lynn2020hierarchical, zhang2023psyattention}, and graph neural networks \cite{yang2021psycholinguistic, yang2023orders, zhu2024enhancing} have been applied to personality detection.
However, despite significant advancements, a major limitation of existing models is their reliance on a uniform text representation approach.

\section{Methodology}
Before introducing MvP, we first clarify the problem formulation of personality detection.
In the personality detection task, each post sample belonging to a certain user is analyzed to capture the user's personality trait in each dimension. 
Therefore, this task can be formulated as a multi-document, multi-label classification task \cite{lynn2020hierarchical, yang2023orders}. 
Formally, given a set of posts $P = \{p_1, p_2, ... p_N\}$ from a certain user, where $p_i = \{w_i^1, w_i^2, ..., w_i^L\}$ represents the $i$-th post with $L$ tokens, the personality detection model need to predict the user's personality traits $Y = \{y_1, y_2, ..., y_T\}$, where each $y_t \in \{0, 1\}$, and $T$ is the number of personality traits. For instance, $T = 4$ in the MBTI taxonomy and $T = 5$ in the Big-Five taxonomy. In this paper, we use the MBTI datasets for validation, hence $T = 4$.

\begin{figure*}[!ht]
\centering
    \includegraphics[scale=0.6, trim=20 400 0 180, clip]{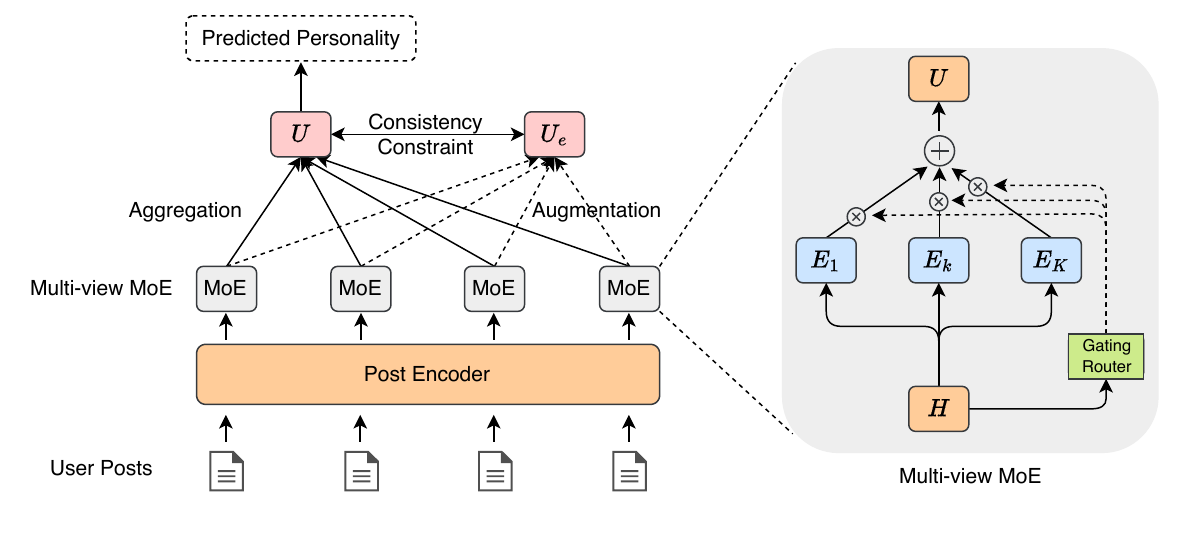}
\caption{The architecture of the MvP model. The left side illustrates the overall framework of MvP, while the right side depicts the structure of Multi-view MoE.}
\label{fig:model}
\end{figure*}

The overall architecture of the proposed MvP is depicted in Figure \ref{fig:model}. MvP initially employs the post encoder consisting of pre-trained language model and word attention mechanism to acquire the fundamental representation of each post. It then proceeds to learn multi-view post representations and aggregate them using a multi-view MoE network based on parameter whitening. Subsequently, MvP learns multi-view generic user representations utilizing user consistency regularization to alleviate semantic conflicts among post representations from different perspectives, thereby acquiring multi-view generalized user representations. Finally, MvP conducts personality detection based on this multi-view generalized user representations. Next, we will provide a detailed description of each component of MvP.

\subsection{Post Encoder}

MvP initially utilizes the powerful contextual representation capability of Pretrained Language Models (PLM) to acquire the fundamental representation of each post. Specifically, MvP employs Equation (\ref{eq:plm}) to obtain the fundamental post representation:

\begin{equation} \small
	\tilde{h}_i = f(p_i) \in \mathbb{R}^{L \times d_w},
\label{eq:plm}
\end{equation}

where $p_i=\{w_i^1, w_i^2, ..., w_i^L\}$ represents the $i$-th post of the user, $f(\cdot)$ represents the function for acquiring the representation of posts in the final hidden states of PLM, and $d_w$ represents the feature dimension of PLM.

As not all words in the same post contribute equally to the prediction of personality, an ideal representation of user post should pay particular attention to personality-revealing portions of a user’s post. 
Thus, MvP introduces word attention to emphasize personality meaningful words of the posts and aggregates the representations of those informative words to form a post representation. Formally,

\begin{equation} \small
\begin{aligned}
p_{i}^l & =\tanh \left(W_{w} \tilde{h}_{i}^l+b_{w}\right) \\
\alpha_{i}^l & =\frac{\exp \left(p_{i}^{l \top} p_{w}\right)}{\sum_{l} \exp \left(p_{i}^{l \top} p_{w}\right)} \\
h_{i} & =\sum_{L} \alpha_{i}^l \tilde{h}_{i}^l .
\end{aligned}
\end{equation}

where $p_w$ is a learnable context vector, $b_w$ is a bias term, and $\alpha_i^l$ is a normalized attention weight for $p_i^l$. $h_i$ is thus a weighted combination of the hidden states representing $\{w_i^1, w_i^2, ..., w_i^L\}$.
Finally, the post encoder yields the representations of all user posts: $H=\{h_1, h_2, ..., h_N\}$.

\subsection{Multi-view Mixture-of-Expert}

Multi-view MoE consists of a group of $K$ Parameter Whitening (PW) based expert networks $E_1$, $E_2$, ..., $E_K$, each catering to specific perspectives, and a gate router  $G$ that outputs a sparse $K$-dimensional vector. The overall structure of Multi-view MoE is depicted on the right side of Figure \ref{fig:model}.

While it's feasible to utilize the "Post Encoder" directly as an expert network to obtain semantic representations from multiple perspectives, the PLM itself contains a large number of parameters, and there's a degree of anisotropy in the textual representations obtained directly by PLM \cite{gao2021simcse, li2020sentence}. This issue exacerbates when acquiring representations from different perspectives simultaneously. 

Hence, inspired by whitening methods \cite{huang2021whiteningbert, hou2022towards}, MvP's Multi-view MoE utilizes parameter whitening, which contains only a negligible number of parameters, to transform original post representations into isotropic semantic representations. This enhances the ability of each expert to represent information from different perspectives, thereby improving MvP's personality detection effectiveness.

Unlike the conventional whitening method that employs preset mean and variance, Multi-view MoE incorporates learnable parameters in the whitening transform to enhance generalizability across different views. Formally, MvP's Parameter Whitening (PW) is represented as follows in Equation (\ref{eq:PW}):

\begin{equation} \small
	PW({h}_{i})=\left(h_{i}-b\right) \cdot W_{1},
\label{eq:PW}
\end{equation}
where $b \in \mathbb{R}^{d_w}$ and $W_1 \in \mathbb{R}^{d_w \times d_v}$ are learnable parameters. The post representations learned through parameter whitening can alleviate the issue of anisotropy to some extent, which is beneficial for learning multi-view post representations.

After obtaining the post representation for the $k$-th view, each expert applies an average aggregator to obtain the user representation for the $k$-th view:

\begin{equation} \small
	u_k = \frac{1}{N} \sum_{i=1}^{N} PW_{k}(h_i)
\label{eq:aggregation}
\end{equation}

We can abstract the $k$-th expert defined by Equation (\ref{eq:PW}-\ref{eq:aggregation}) as a function:
\begin{equation} \small
	u_k = E_k(H)
\end{equation}

Then the computation process of the Multi-view MoE module can be expressed by Equation (\ref{eq:MoE_layer}):
\begin{equation} \small
	U = \sum_{k=1}^{K} G_k(H) \cdot E_k(H)
\label{eq:MoE_layer}
\end{equation}
where $G_k(\cdot)$ is the weight of $k$-th gating router, and its calculation is defined by Equation (\ref{eq:gate}):
\begin{equation} \small
\begin{aligned}
	G(H) &= \text{Softmax}\left(h \cdot W_{2} + \delta \right) , \\
	h &= \frac{1}{N} \sum_{i=1}^{N} h_i.
\end{aligned}
\label{eq:gate}
\end{equation}

In Equation (\ref{eq:gate}), MoE utilizes the user post representation $H$ as the input of the gating route $G(\cdot)$. 
The learnable parameter matrices $W_2, W_3 \in \mathbb{R}^{d_w \times G}$ allow for adaptive tuning of the expert weights $G(\cdot) \in \mathbb{R}^G$. 
Furthermore, to balance the expert load, the noise term $\delta$ is introduced:
\begin{equation} \small
\begin{aligned}
	\delta &= \text{Softplus}\left(h \cdot W_{3}\right) \cdot \varepsilon,
\end{aligned}
\end{equation}
where $\varepsilon$ is randomly generated Gaussian noise, and $\text{Softplus}(\cdot)$ is the Softplus activation function.

Utilizing the Multi-view MoE, MvP learns multiple expert networks for different views simultaneously based on parameter whitening, enhancing the post and user representations from different perspectives. 
With post representations from each perspective enhanced by parameter wightening, the Multi-view MoE aggregates post representations within each perspective, obtaining a multi-view user representation. 
Moreover, leveraging the learnable gating router, MvP's MoE adaptively establishes the relevance of different perspectives, facilitating the fusion and adaptation of information from diverse perspectives and obtaining a multi-view enhanced user representation $U$.

\subsection{User Consistency Regularization}

As mentioned earlier, intereference and conflicts may arise among different views. To mitigate this issue, MvP introduces User Consistency Regularization, aiming to learn a multi-view generalized user representation for personality detection. Drawing inspiration from \cite{li2020sentence} and \cite{wu2021r}, MvP ensures the consistency of user representation induced by the Dropout technique \cite{hinton2012improving}.

During training, each user sample undergoes two forward passes, with each pass being processed by a different sub-model by randomly dropping out some hidden units. The user representations $U$  and $U_e$ obtained from the two sub-models are mutually augmentation of each other. 

MvP ensures that the two personality distributions $p(\hat{y}_t|U)$ and $p(\hat{y}_t|U_e)$ for the same user, calculated by the augmented user representation pair, are consistent with each other by minimizing the bidirectional Kullback-Leibler (KL) divergence between the two distributions as Equation (\ref{eq:loss_ucr}).

\begin{equation} \small
\begin{aligned}
	\mathcal{L}_{ucr} &= \frac{1}{2}\left(\mathcal{D}_{K L}\left(p(\hat{y}_t|U) \| p(\hat{y}_t|U_e))+\mathcal{D}_{K L}(p(\hat{y}_t|U_e) \| p(\hat{y}_t|U)\right)\right) \\
	p(\hat y_t) &= \text{Softmax}(UW_u + b_u)
\end{aligned}
\label{eq:loss_ucr}
\end{equation}

\subsection{Objective Function}

To minimize the difference between the predicted probability distributions and the actual personality trait labels, we use the binary cross-entropy loss as the personality detection loss function:
\begin{equation} \small
	\mathcal{L}=\sum_{t=1}^{T} y_{t} \log \left(p(\hat{y}_{t})\right)+\left(1-y_{t}\right) \log \left(1-p(\hat{y}_{t})\right).
\label{eq:loss_det}
\end{equation}

We adopt a multi-task joint learning strategy, where we employ the auxiliary User Consistency Regularization to assist in learning supervised personality detection during training, and the two tasks share the same parameters. The training objective is to minimize both the cross-entropy loss $\mathcal{L}_{det}$ and KL loss $\mathcal{L}_{ucr}$, corresponding to the supervised personality detection and the self-supervised user consistency constraint, respectively. Specifically, the objective function of MvP is defined as follows:

\begin{equation} \small
	\mathcal{L} = \mathcal{L}_{det} + \lambda \mathcal{L}_{ucr},
\label{eq:loss}
\end{equation}
where $\lambda$ is a hyperparameter used to control the weight of the user consistency regularization loss $\mathcal{L}_{ucr}$.

\section{Experimental Settings}

\subsection{Datasets}

We have chosen two widely used personality detection datasets, Kaggle and Pandora MBTI, for our research.
The MBTI personality model classifies personality types into four characteristics: \textit{Extroversion} vs. \textit{Introversion} (E/I), \textit{Sensing} vs. \textit{Intuition} (S/N), \textit{Thinking} vs. \textit{Feeling} (T/F), and \textit{Judging} vs. \textit{Perception} (J/P).

The Kaggle dataset\footnote{\url{https://www.kaggle.com/datasets/datasnaek/mbti-type}} consists of 8675 users, each contributing approximately 45-50 posts. 
The Pandora dataset\footnote{\url{https://psy.takelab.fer.hr/datasets/all/}} contains dozens to hundreds of posts for each of the 9067 users.
Because both personality detection datasets exhibit severe distribution imbalances across different personality characteristics, we employed the Macro-F1 metric to better evaluate performance on each characteristic.
The average Macro-F1 of all personality characteristics was used to assess overall performance.
The datasets were shuffled and divided into a 6:2:2 split for training, validation, and testing, respectively. Consistent with prior studies \cite{yang2023orders}, we removed all words matching any personality label to prevent information leaks. 

\subsection{Baseline Methods}

We compare MvP with the following state-of-the-art models in the personality detection task. 
\textbf{(1) SVM} \cite{cui2017survey} \textbf{and XGBoost} \cite{tadesse2018personality}: Classify personality characteristics based on TF-IDF features from concatenated user-generated posts.
\textbf{(2) BiLSTM} \cite{tandera2017personality}: Encodes each post using a Bi-LSTM with GloVe word embeddings and averages post representations for the user representation.
\textbf{(3) AttRCNN} \cite{xue2018deep}: Uses a CNN-based aggregator to obtain user representation, incorporating psycholinguistic knowledge with LIWC features.
\textbf{(4) BERT} \cite{keh2019myers}: Fine-tuned BERT encodes each post, and the user representation is derived by mean pooling over post representations.
\textbf{(5) SN + Attn} \cite{lynn2020hierarchical}: Hierarchical attention network with GRU for word-level attention and another GRU for post-level attention to generate the user representation.
\textbf{(6) TrigNet} \cite{yang2021psycholinguistic}: Constructs a tripartite graph (post, word, category) and uses a graph neural network to aggregate post information.
\textbf{(7) D-DGCN} \cite{yang2023orders}: Dynamic graph neural network learning correlations within user posts and aggregates them based on the learned graph structure.
Notably, D-DGCN employed a domain-adaptive pretrained BERT as its backbone. To ensure equitable experimental comparison, we adopt D-GCN as a baseline with the original pretrained BERT.

\subsection{Implementation Details}

All models were implemented using PyTorch and trained on GeForce RTX 3090 GPUs. We utilized the pretrained BERT from Hugging Face\footnote{\url{https://huggingface.co/google-bert/bert-base-cased}} as the pretrained language model. 
The balancing hyperparameter $\lambda$ in Equation \ref{eq:loss} is set to 1.0, and the number of multi-view experts is set to 6 ($K=6$). All hidden layer dimensions in MvP were set to 768.
We established specific parameters for the Kaggle and Pandora datasets, limiting the maximum number of posts to 50 for Kaggle ($N=50$) and 100 for Pandora ($N=100$), with a standardized maximum post length of 70 ($L=70$) characters.
Training was performed using the Adam optimizer \cite{kingma2014adam} with a learning rate of 2e-5 for the pretrained BERT in the Post Encoder and 2e-3 for other components. The mini-batch size was 32, and training continued for up to 10 epochs. The model with the best validation performance was chosen for testing.

\section{Overall Results}

\begin{table*}[!ht]
\footnotesize
\centering
\caption{Overall Macro-F1 (\%) results of MvP and baseline Models on the Kaggle dataset and Pandora MBTI datasets, where the best results are shown in bold, and the suboptimal results are underlined. Statistical significance of pairwise differences for MvP against the best baseline is determined by the t-test (p< 0.05).}
\begin{tabular}{lcccccccccc}
\toprule[1.5pt]
\multirow{2}{*}{\textbf{Approch}} & \multicolumn{5}{c}{\textbf{Kaggle}} & \multicolumn{5}{c}{\textbf{Pandora}} \\
\cmidrule[1pt](lr){2-6}
\cmidrule[1pt](lr){7-11}
& \textbf{E/I} & \textbf{S/N} & \textbf{T/F} & \textbf{J/P} & \textbf{Average} & \textbf{E/I} & \textbf{S/N} & \textbf{T/F} & \textbf{J/P} & \textbf{Average}\\
\midrule[1pt]
SVM & 53.34 & 47.75 & 76.72 & 63.03 & 60.21 & 44.74 & 46.92 & 64.62 & 56.32 & 53.15 \\
XGBoost & 56.67 & 52.85 & 75.42 & 65.94 & 62.72 & 45.99 & 48.93 & 63.51 & 55.55 & 53.50 \\
BiLSTM & 57.82 & 57.87 & 69.97 & 57.01 & 60.67 & 48.01 & 52.01 & 63.48 & 56.12 & 54.91 \\ 
AttRCNN & 59.74 & 64.08 & 78.77 & 66.44 & 67.25 & 48.55 & 56.19 & 64.39 & 57.26 & 56.60 \\
BERT & 65.02 & 59.56 & 78.45 & 65.54 & 67.14 & 56.60 & 48.71 & 64.70 & 56.07 & 56.52 \\
SN + Attn & 62.34 & 57.08 & 69.26 & 63.09 & 62.94 & 54.60 & 49.19 & 61.82 & 53.64 & 54.81 \\
TrigNet & \textbf{69.54} & \uline{67.17} & 79.06 & \uline{67.69} & \uline{70.86} & 56.69 & 55.57 & 66.38 & 57.27 & 58.98 \\

D-DGCN & 67.37 & 64.47 & \uline{80.51} & 66.10 & 69.61 & \uline{58.28} & \uline{55.88} & \uline{68.50} & \uline{57.72} & \uline{60.10} \\

\midrule[1pt]
MvP(Ours) & \uline{67.68} & \textbf{69.89} & \textbf{80.99} & \textbf{68.32} & \textbf{71.72} & \textbf{60.08} & \textbf{56.99} & \textbf{69.12} & \textbf{61.19} & \textbf{61.85}	\\
\bottomrule[1.5pt]
\end{tabular}
\label{tb:Overall Results MBTI}
\end{table*}

The results of the comparison between MvP and the baseline models are shown in Table \ref{tb:Overall Results MBTI}, with the optimal results bolded and the suboptimal results underlined. The results indicate that the average Macro-F1 scores of MvP on the two datasets are 71.72\% and 61.85\%, respectively, surpassing all baseline models.
MvP performed exceptionally well across each personality dimension as well. 
On the Kaggle dataset, MvP achieved the highest Macro-F1 values in the S/N, T/F and J/P dimensions, while performing second only to TrigNet in the E/I dimension. On the Pandora dataset, MvP attained the highest Macro-F1 values across all personality trait dimensions.

Overall, MvP demonstrates excellent capability in personality detection, successfully validating its effectiveness in this task. 
MvP's outstanding performance can be attributed to several key factors.
Firstly, MvP's multi-view MoE effectively models user posts from various perspectives and mitigates the anisotropy problem present in post encoding. 
Additionally, MvP employs the user consistency regularization to address conflicts between semantic representations from different views, resulting in a multi-view generalized user representation that ultimately enhances the model's effectiveness in personality detection.

\section{Detailed Analysis}

\subsection{Ablation Study}

\begin{table}[!ht]
\small
\centering 
\caption{Results of various variants of MvP model in Macro-F1 on Kaggle dataset.}
\begin{tabular}{lccccc}
\toprule[1.5pt]
\textbf{Approach} & \textbf{E/I} & \textbf{S/N} & \textbf{T/F} & \textbf{J/P} & \textbf{Average}\\
\midrule[1pt]
MvP & 67.68 & 69.89 & 80.99 & 68.32 & 71.72              
\\
w/o MoE & 67.60	& 67.39	& 80.67	& 67.23	& 70.72 \\
w/o UCR & 66.54 & 69.03 & 80.19 & 65.92 & 70.42 \\
\bottomrule[1.5pt]
\end{tabular}
\label{tb:ablation}
\end{table}

To analyze the impact of each key module in MvP more thoroughly, a series of ablation experiments were conducted on the Kaggle dataset, and the results are shown in Table \ref{tb:ablation}.
As shown in Table \ref{tb:ablation}, removing any part of the model resulted in varying degrees of performance degradation. Specifically, removing the Multi-view MoE resulted in a significant performance drop, demonstrating the importance of modeling user posts from multiple perspectives and the effectiveness of the Multi-view MoE in this regard. 
Similarly, a notable decrease in model performance was observed after removing the user consistency regularization, confirming that user consistency regularization effectively mitigates conflicts in the representation of posts from different perspectives.

\subsection{Effect of Number of Views}
\label{sec:effect of number of views}

\begin{figure}[!ht]
\centering
    \includegraphics[scale=0.22]{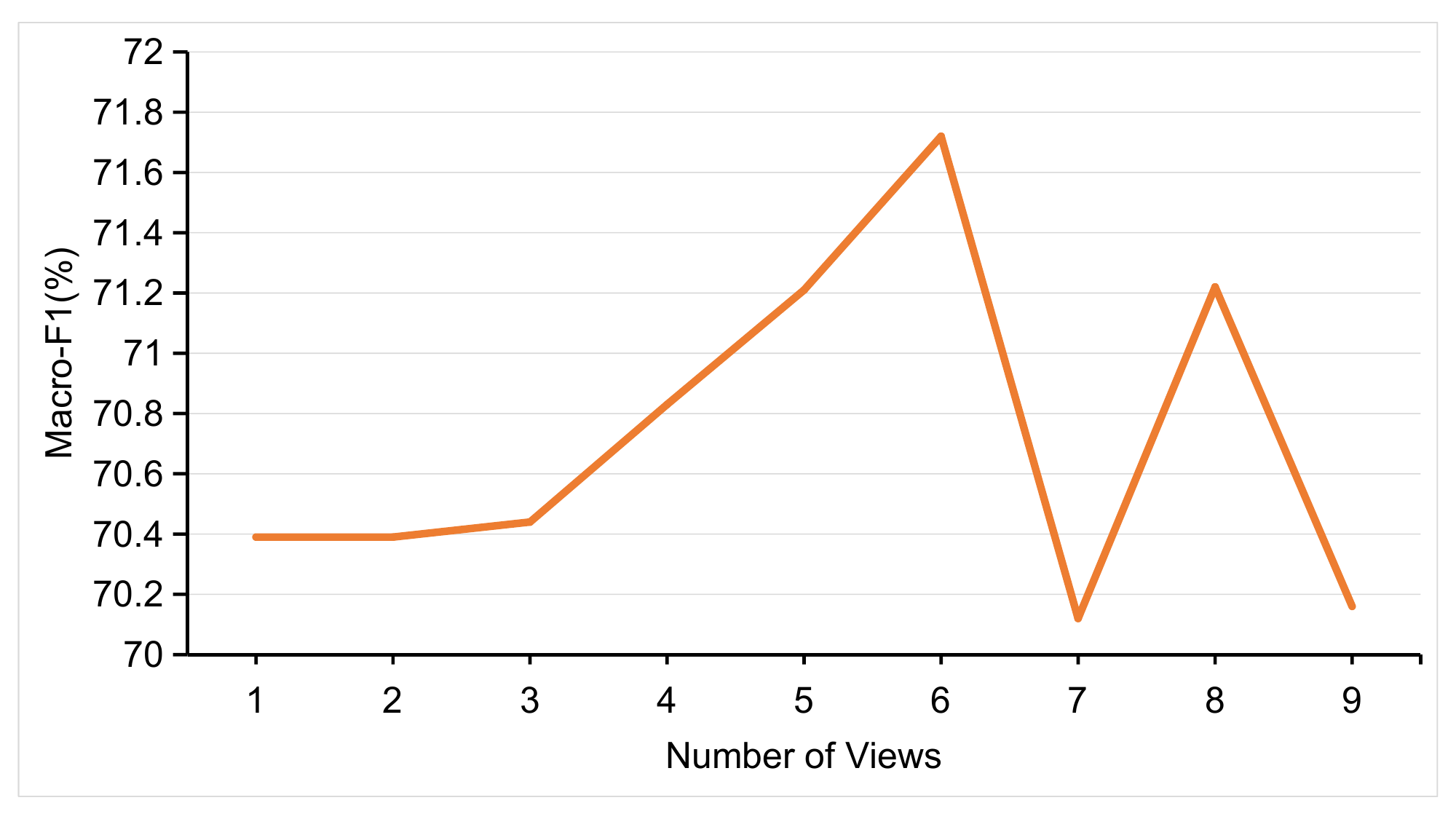}
\caption{Performance curves for different number of views.}
\label{fig:effect of the number of views}
\end{figure}

To study the effect of the number of views on MvP's personality detection performance, we conducted experiments on the Kaggle dataset by adjusting the number of experts for different views. 
The results, shown in Figure \ref{fig:effect of the number of views}, indicate that performance improves as the number of views increases, peaking at 6 views. Beyond this point, performance declines. 
The initial improvement is due to more views capturing richer post and user representations, demonstrating the effectiveness of modeling posts from different perspectives. The decline with excessive views is due to increased conflicts between representations from different views, underscoring the importance of mitigating these conflicts. MvP addresses this with user consistency regularization, but with too many views, conflicts become inevitable, leading to reduced performance.

\subsection{Effect of Trade-off Parameter}

\begin{figure}[!ht]
\centering
    \includegraphics[scale=0.22]{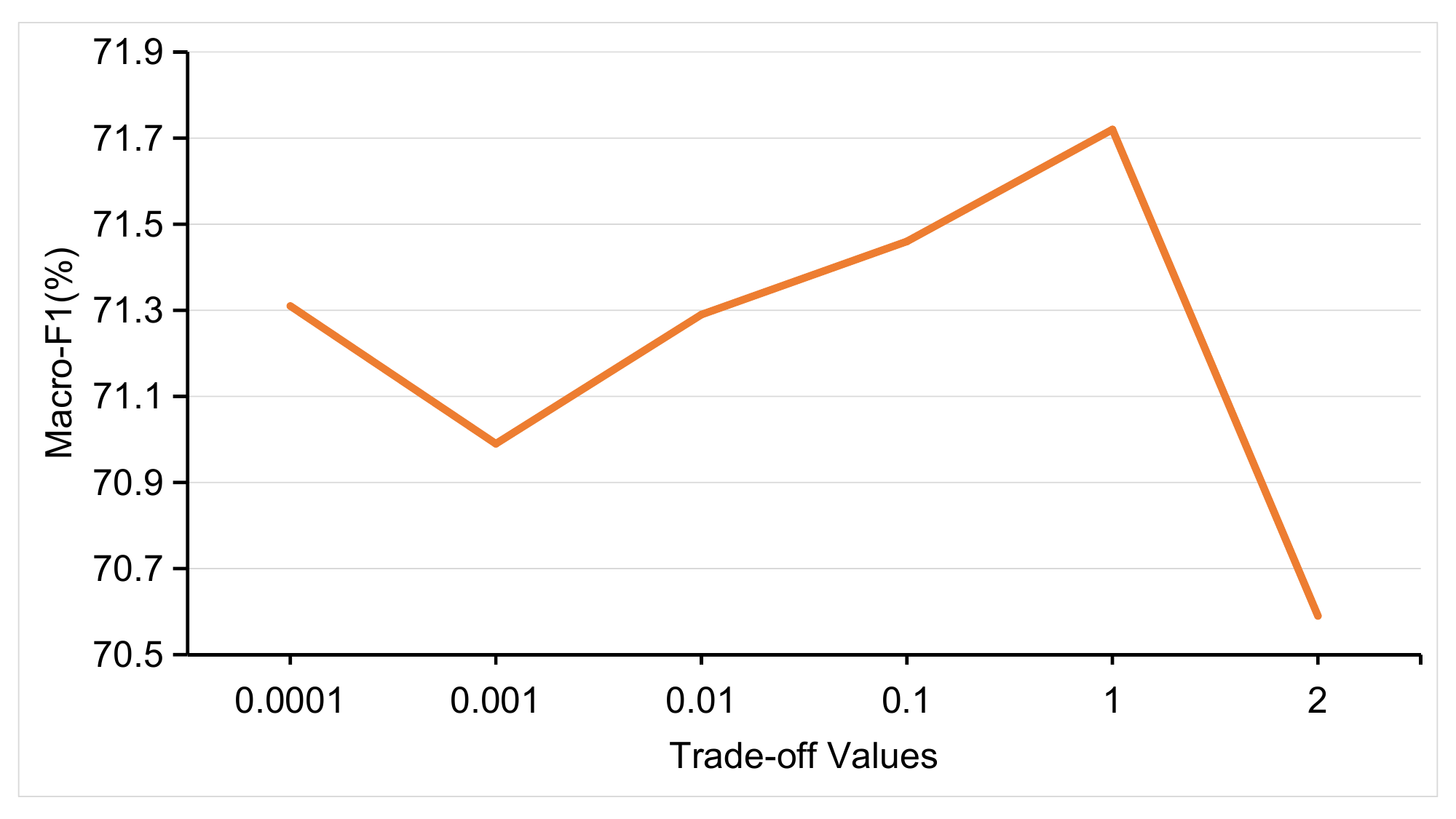}
\caption{Performance curves for different trade-off parameters.}
\label{fig:effect of trade-off parameter}
\end{figure}

To investigate the effect of different trade-off parameters $\lambda$ in Equation (\ref{eq:loss}) on model performance, we set various equilibrium parameters for MvP and conducted experiments on the Kaggle dataset.
Figure \ref{fig:effect of trade-off parameter} illustrates how the performance of MvP changes as the trade-off parameter $\lambda$ increases.

From Figure \ref{fig:effect of trade-off parameter}, it can be observed that performance gradually increases as the balancing parameter $\lambda$ increases, but starts to decrease when $\lambda$ exceeds 1.0. 
This is because a larger value of $\lambda$ enhances the impact of the user consistency regularization, helping to mitigate conflicts and over-characterization in multi-view post representations. 
However, if the $\lambda$ value is too large, it can lead to performance degradation as excessive user contrast learning can interfere with the primary personality detection task.

\section{Conclusion}

In this paper, we introduced MvP, a model designed to effectively analyze user posts from multiple perspectives, acknowledging the diverse aspects of user-generated content to enhance personality detection.
MvP introduces a Multi-view Mixture-of-Experts (MoE) approach to learn post representations from various perspectives.
To address semantic conflicts between these perspectives, MvP incorporates User Consistency Regularization, resulting in a multi-perspective generalized user representation for improved personality detection.
Our experiments provided empirical evidence for the efficacy of MvP in personality detection and underscore the benefits of automatically modeling user posts from various perspectives.

\bibliographystyle{splncs04}
\bibliography{bibliography}

\end{document}